\documentclass[11pt,a4paper]{article}
\usepackage[nohyperref]{eacl2021}
\usepackage[T1]{fontenc}
\usepackage{times}
\usepackage{latexsym}
\usepackage{url}

\usepackage{linguex}
\usepackage{booktabs}
\usepackage{graphicx}
\usepackage{enumitem}

\usepackage{microtype}

\setlist{nosep}

\aclfinalcopy 

\title{Unsupervised Discovery of Unaccusative and Unergative Verbs}

\author{
Sharid Loáiciga$^{1}$ \hspace*{5mm} Luca Bevacqua$^{2}$ \hspace*{5mm} Christian Hardmeier$^{3,4}$ \\
$^1$CLASP, Department of Philosophy, Linguistics \& Theory of Science, University of Gothenburg\\
$^2$School of Philosophy, Psychology \& Language Sciences, University of Edinburgh \\
$^3$Department of Linguistics and Philology, Uppsala University\\
$^4$Department of Computer Science, IT University of Copenhagen\\
{\tt sharid.loaiciga@gu.se \hspace*{5mm} luca.bev@hotmail.it}\\
{\tt chrha@itu.dk}}

\date{}

\begin{document}
\maketitle
\begin{abstract}

We present an unsupervised method to detect English unergative and unaccusative verbs. These categories allow us to identify verbs participating in the causative-inchoative alternation without knowing the semantic roles of the verb. The method is based on the generation of intransitive sentence variants of candidate verbs and probing a language model. We obtained results on par with similar approaches, with the added benefit of not relying on annotated resources.

\end{abstract}

\section{Introduction}

As NLP systems push towards Natural Language Understanding, their ability to grasp verb meaning is central. In this paper we present an unsupervised method to detect English unergative and unaccusative verbs. Within the wider category of intransitive verbs, these subgroups show differences in their behaviour, such as (in different languages) auxiliary selection, passivisation, cliticisation and causative-inchoative alternation. These differences are due to the different semantic roles of the subject of unergative verbs, which shares its agentivity with subjects in transitive frames, and that of unaccusative verbs, more similar to the stereotypically patient- or theme-like objects in transitive frames.

Moreover, these categories relate to the causative-inchoative alternation \citep{Haspelmath1993} in which unaccusative verbs can express a same event with either agent and patient \ref{ex:balloon-tran} or with a patient only \ref{ex:balloon-intr}, corresponding respectively to a causative and inchoative interpretation of the event:

\ex. \a. Hannah popped the balloon.\label{ex:balloon-tran}
\b. The balloon popped.\label{ex:balloon-intr}

The main property of the causative-inchoative alternation is the ability of the patient to be promoted from object to subject, and most theoretical accounts focus on this transitive to intransitive frame change. Once the event is in its intransitive form, however, it is syntactically indistinguishable from any other intransitive verb construction, and in order to categorize the verb one must know the specific semantic roles of its arguments. 

This phenomenon is particularly hard to mine because the different realisations are not disambiguated by the context but by the arguments of the verb, and the frequency of the constructions themselves is not an indication of their acceptability. Furthermore, the meaning of the sentence remains virtually the same.\footnote{To the point where in the intransitive version of the alternation, an agent can be surmised as implicit, rather than simply not existing.}  However, picking up this categorisation is important not only for reasons involving the appropriateness of different types of subjects for each verb, but also because it has been shown to influence coreference patterns \citep{loaiciga-etal-2018}. Nevertheless, efforts to discover verbs participating in the alternation using automatic methods are very limited.


Here we focus on the unaccusative vs unergative distinction as it allows us to disambiguate the verbs based on their intransitive frames. While the subjects of unaccusatives are patients \ref{ex:balloon-intr}, subjects of unergatives are agents \ref{ex:sleep-intr}. The assumption is that alternating verbs belong to the unaccusative category and we can discover them by separating them from the unergative category.\footnote{Note that this assumption does not exclude the exceptional cases of unaccusatives that do not alternate, e.g., \textit{to die}, in \textit{The cactus died a slow death}. Examples of this kind include an internal (or cognate) object, a noun that can be used as the object of an ordinarily intransitive verb by virtue of their semantic similarity.}

\ex. \a. Hannah slept. \label{ex:sleep-intr}

The key in using these categories is that, without knowing the semantic roles of the verb, we can measure how well a noun fits the subjective position using a large corpus. Our method relies on language modeling to do just this. We investigate the effect of different language models on the task of identifying unaccusatives vs unergatives by testing the intransitive frames of a large quantity of verbs. 




\begin{table*}[h]\centering
\resizebox{\linewidth}{!}{
\begin{tabular}{l*{19}{c}}
\toprule
 & &\multicolumn{6}{c}{\textbf{n-gram small}} & \multicolumn{6}{c}{\textbf{n-gram large}} & \multicolumn{6}{c}{\textbf{GPT-2}} \\
 &  &   \multicolumn{3}{c}{Unacc.} & \multicolumn{3}{c}{Unerg.} & \multicolumn{3}{c}{Unacc.} & \multicolumn{3}{c}{Unerg.} & \multicolumn{3}{c}{Unacc.} & \multicolumn{3}{c}{Unerg.} \\
  & Verbs &   P & R & F1& P & R & F1& P & R & F1& P & R& F1& P & R & F1& P & R & F1\\
 \midrule
\textbf{Constructed}&20&0.20&0.25& 0.22 & 0.40 & 0.33 & 0.36 & 0.80  & 0.61 & 0.69 &0.50  & 0.71 & 0.58  & 0.62 &0.80&0.70 & 0.71 & 0.50 & 0.59\\
 \bottomrule
\end{tabular}}
\caption{Precision, recall and F-scores using a manually curated data set. We present results using the present tense.}\label{tab:resultsgold}
\end{table*}

\section{Related Work}

\citeauthor{Levin1993}'s~\citeyearpar{Levin1993} seminal work on verb alternations remains the most comprehensive collection of alternating verbs for English. Other collections, for instance Framenet \citep{Collin-etal-FRAMENET} also specify if a verb allows the alternation based on \citeauthor{Levin1993}. Typological work for other languages exists in the linguistics literature \citep{Haspelmath1993,Shaefer_2009_alternation}, but large collections are practically nonexistent.\footnote{The World Atlas of Transitivity Pairs \citep{WATP}, for instance, includes many languages but only a few verbs.} 

Building on \citeauthor{Haspelmath1993}'s theory, \citet{Samardzic-2014-thesis} estimates a Spontaneity score based on the ratio of a verb's transitive to intransitive occurrence. In this scale, verbs are ranked according to the degree to which they are non-agentive. In other words, verbs without an explicit agent causing the event are more likely to participate in the causative alternation.  \citet{Samardzic_Merlo_2018} report between 61\% and 85\% agreement between their model and theoretical classifications.  

\citet{kann-etal-2019-verb} rely on \citeauthor{Levin1993}'s work to create synthetic data and build classifiers able to discriminate between several alternations. Their data sets are built using proper names as subjects and common nouns as objects, and they focus on the transitive to intransitive construction case. In this paper we rely on intransitive constructions exclusively, as will be explained below in Section \ref{sec:probingsents}. In addition, we use their FAVA data set for evaluation in Section \ref{sec:evaluation}. 

Our method is most similar to the RNN-based method reported by \citet{seyffarth-2019-identifying}. Contrary to our work which queries language models with inflected sentences, \citeauthor{seyffarth-2019-identifying} uses an RNN to score artificially created transitive and intransitive argument sequences of the type \textit{invite-Pat-Kim} vs \textit{invite-Pat}. Using Framenet as gold standard, they report an accuracy of 66\% on all verbs. 


\section{Generating Probing Sentences}\label{sec:probingsents}


To generate probing sentences, we start by extracting transitive verb frames
from a corpus. We start with a version of Europarl \citep{koehn2005epc} with dependency annotations automatically generated with the
Stanza parser \citep{qi2020stanza} trained on Universal Dependencies v2.5. In these parses,
we extract lemmas of transitive verbs with the head nouns of their subject and
object noun phrases. In particular, we search for words tagged with a
\textsc{verb} part-of-speech tag having a dependent in an \textit{obj} relation,
its direct object. We find the subject of these verbs either as their
\textit{nsubj} children, or, if they are the head of a relative clause, as their
head, to which they are linked with an \textit{acl:relcl} dependency arc. We
extract an example, consisting of a triplet of \textit{(subject, verb, object)}
lemmas, if both the subject and the object are nouns. We assume
that the \emph{subjects} of transitive verbs are typically in an
\emph{agent}-like relation with the predicate, so we treat them as agent
candidates, whereas the \emph{objects} are more likely to be in a
\emph{patient}-like relation and are used as patient candidates.


In a next step, we expand the set of agent and patient nouns by using it to seed
the lookup of semantically related words in a non-contextual word embedding
space generated by \citet{pennington2014glove}. We start with the pretrained
300-dimensional \emph{glove-wiki-gigaword-300} model from the \emph{gensim}
library \citep{rehurek_lrec}. First, we filter the vocabulary of the GloVe model
so that it only contains nouns. Then, we expand the word sets according to the
following procedure:
\begin{enumerate}
\item Let $V$ be the original vocabulary of the embeddings space, and $S$ and
$O$ be the sets of words observed in subject and object position of transitive
verb frames, respectively.
\item Disjoint sets of seed words are created as $S'=V\cap S\setminus O$ and
$O'=V\cap O\setminus S$.
\item We proceed as follows to create expanded sets $S^{+}$ and $O^{+}$ from
$S'$ and $O'$, respectively:
\begin{enumerate}
\item We draw 20 samples of 10 items from the seed word list, $S'$ or $O'$.
\item For each sample, we find the 50 nearest neighbours in the embedding space
using the \textsc{3CosMul} similarity metric of
\citet{levy-goldberg-2014-linguistic}. The union of these 20 sets of nearest
neighbours forms the expansion candidates.
\item Disjoint sets $S^{+}$ and $O^{+}$ are created by taking the 30
highest-scoring expansion candidates generated from $S'$ and $O'$ respectively,
but ignoring items that occur in the subject and object expansion candidates of
the same verb.
\end{enumerate}
\end{enumerate}

Probing sentences are generated by inserting the items of the agent-like and
patient-like expanded sets into templates of the form
\begin{quote}
<s> The \textsc{noun} \textsc{verb}s . </s>
\end{quote}
These sentences are then scored with the language model, resulting in a
probability for each \textsc{noun-verb} pair. Finally, we sum up the probability
of the agent-like and patient-like nouns separately. We classify a verb as
unaccusative if the total probability of the sentences with patient-like fillers
exceeds that of the sentence with agent-like fillers, and as unergative
otherwise.


\section{Evaluation}\label{sec:evaluation}

We test our method on three data sets. The first is a small set with manually
constructed examples containing 10 verbs of each category. The second is the
subset of causative-inchoative alternating verbs from the FAVA data set. This
means that it only contains verbs of the unaccusative class. The FAVA data set
includes both the transitive and intransitive variants of each verb. However,
since each sentence uses a proper name as subject and we do not consider proper
names in our pattern, ``<s> The \textsc{noun} \textsc{verb}s . </s>'', we just retain the verbs. The third evaluation data set is composed of the subset of FrameNet verbs annotated as unaccusative or unergative verbs. We randomly sample 50\% of the verbs each time we query the language models. For the first two data sets, we present results both with the manually curated sentences and with generated sentences (according to Section \ref{sec:probingsents}). Since we do not have manually curated sentences for the FrameNet data set, we only provide results with generated sentences.

In terms of language models, we experiment with two n-gram models and the neural
model GPT-2 \citep{radford2019language-gpt2}. The first two are 5-grams models
with modified Kneser-Ney smoothing \citep{Chen:1998,Kneser:1995}
trained using KenLM \citep{heafield-etal-2013-scalable}. They have different
sizes: the small one is trained on the News Commentary portion of the
shared-task training data for the Conference on Machine Translation (WMT) 2016
\citep{bojar-etal-2016-findings}, while the large one is trained using the
entirety of the monolingual English training data of the WMT 2016 news
translation shared task. 

For the neural model, we normalised the sentence log probabilities produced by GPT-2 by sentence length and lexical frequency following \citet{lau-etal-2015-unsupervised}.\footnote{
The normalisation LP-div is defined as $-\frac{\log P_m(\xi)}{\log P_u(\xi)}$,
while the SLOR (syntactic log-odds ratio) is defined as $\frac{\log P_m(\xi) -
\log P_u(\xi)}{|\xi|}$, where $P_m(\xi)$ is the probability of the sentence given
by the model and $P_u(\xi$) is the unigram probability of the sentence.} Since some of the verb variants risked to be infrequent but grammatical, we expected the normalisation to provide a better estimate of the suitability of each query. We estimated an unigram language model on Europarl for the normalisation.

\section{Results}

\begin{table*}[t]
\resizebox{\linewidth}{!}{
\begin{tabular}{l*{19}{c}}
\toprule
& &\multicolumn{6}{c}{GPT-2 unnormalized} & \multicolumn{6}{c}{GPT-2 norm-LP-div} &  \multicolumn{6}{c}{GPT-2 norm-SLOR} \\
 &  &  \multicolumn{3}{c}{Unacc.} & \multicolumn{3}{c}{Unerg.} & \multicolumn{3}{c}{Unacc.} & \multicolumn{3}{c}{Unerg.} & \multicolumn{3}{c}{Unacc.} & \multicolumn{3}{c}{Unerg.} \\
  & Verbs &  P & R & F1&  P & R& F1 &  P & R & F1& P & R &F1 &  P & R&F1 &  P & R&F1 \\
 \midrule
 \textbf{Constructed} & 20 & & & & & & & & & & & & & & & & & & \\
Expanded\_EP& &0.69& 0.90& 0.78 & 0.86 & 0.60 & 0.71&0.62 & 0.80 &0.70 & 0.71 &0.50& 0.59& 0.62&0.80&0.70& 0.71 &0.50& 0.59 \\

Expanded\_Lefff& &0.69&0.90&0.78 &0.86 & 0.60&0.71&0.54 &0.70&0.61 & 0.57&0.40&0.47 &0.54 &0.70&0.61 &0.57 & 0.40 &0.47\\

\midrule
 \textbf{FAVA}  & 120 & & & & & & & & & & & & & & & & & & \\
Expanded\_EP & &0.69 &0.34  &0.45 &0.56 &0.69 &0.62 & 0.50 &0.32 &0.39 &0.45 &0.45  &0.45 &0.49 &0.32  &0.39& 0.44 &0.43 & 0.43\\

Expanded\_Lefff& &0.70 &0.30& 0.42&0.66 & 0.65&0.65&0.63 &0.31& 0.42 &0.64 & 0.57& 0.60&0.67 &0.31& 0.42 &0.66 &0.61& 0.63\\

\midrule
\textbf{FrameNet} &329 & & & & & & & & & & & & & & & & & & \\
Expanded\_EP & &0.60 &0.15& 0.24 &0.40 &0.15& 0.22 &0.59 &0.15& 0.24 &0.38 &0.14&0.20 &0.59 &0.15& 0.24 &0.38 &0.14& 0.20\\

Expanded\_Lefff& &0.62&0.09&0.16 &0.43 &0.13&0.20 &0.61 &0.13& 0.22&0.43 &0.10&0.16 &0.67 &0.14 &0.23&0.49 &0.12& 0.19 \\
 \bottomrule
\end{tabular}}
\caption{Precision, recall and F-scores for the GPT-2 model on the different evaluation sets. We present results using the present tense. Notation: Expanded\_EP refers to probing sentences generated with any word tagged as \textsc{noun} in English Europarl; Expanded\_Lefff refers to probing sentences generated with all nouns listed in the English dictionary Lefff.}\label{tab:resultsexpanded}
\end{table*}


The difference between the n-gram models confirms that it is possible to differentiate between the classes using intransitive frames only and that a larger model produces better predictions. The results of the neural model, on the other hand, seem to contradict this last statement. 

We know that the n-gram model may do hard back-offs when not recognising a 3-gram ``<noun> <verb> .'', causing the final score to depend only on the prior frequency of the subject. However, it is hard to diagnose what it is that the neural model is reacting to. 

The neural model has a maximal precision of 0.69 in the Constructed set and a precision of 0.60 in the FrameNet set. The comparison of the evaluation on the Constructed set using simple sentences (Table \ref{tab:resultsgold}) with the expanded sentences generated as explained in Section \ref{sec:probingsents} (Table \ref{tab:resultsexpanded}) suggests that the method is robust. We observe an improvement of 0.07 in precision and of 0.1 in recall. Likewise, we observe an improvement of 0.10 in precision for the unaccusative class in the FAVA set. 

We produced expanded sets of potential agents and patients using different corpus sources for the probing sentences. In Table \ref{tab:resultsexpanded}, we report results using any word tagged as \textsc{noun} by Stanza in English Europarl (vocabulary size of 22,923 words), and also probing sentences generated with all nouns listed in the English dictionary Lefff \citep{sagot-2010-lefff} (vocabulary size of 49,932 words). We observe that they produce similar precision results, but the Europarl expansion yields better recall, in reversed correlation with the vocabulary sizes. 

In addition, our method predicts labels for a large number of verbs unattested in the existing gold standards, as reflected on the low recall scores across all settings. Interestingly and in contrast to \citet{lau-etal-2015-unsupervised}, normalising the neural model scores systematically hurts the classification scores. 

Comparisons with similar methods are difficult. We have a close set up to \citet{seyffarth-2019-identifying}, however, while they query their neural model with argument sequences, we probe ours with inflected sentences. In addition, while \citet{kann-etal-2019-verb} report results between 66\% and 85\% accuracy for the inchoative verbs, their task is supervised and only considers verbs known to alternate. 

Finally, we also experimented with computing the probability of the </s> token instead of the probability of the complete sentence. However, this strategy did not improve our results.

\section{Conclusions}

We proposed a method to detect unaccusative vs unergative verbs based on the generation of intransitive sentence frames of candidate verbs. The results with a large language model show moderate success, highlighting that the causative-inchoative alternation is a challenging meaning distinction to detect automatically. Since the method relies primarily on parsed data and language models, it has the potential to be extended to languages where verbal annotated resources are scarce. 

\section*{Acknowledgements}

Christian Hardmeier was supported by the Swedish Research Council under grant 2017-930.

\bibliographystyle{acl_natbib}
\bibliography{eacl2021,anthology}

\end{document}